\documentclass[rn,11pt]{ucl_document} 
\usepackage{graphicx}
\usepackage{latexsym}
\usepackage{hyperref} 
\usepackage{named}

\newlength{\halfwidth}
\newlength{\temp}
\newlength{\tempa}

\begin{document}

\title{Using Genetic Programming to Model Software}

\author{
{W. B. Langdon} 
\and 
{M. Harman}
}

\date{24 June 2013} 
\documentnumber{13/12}

\maketitle

\begin{abstract}
We study a generic program to investigate the scope for
automatically customising it for a vital current task,
which was not considered when it was first written.
In detail,
we show genetic programming (GP) can evolve models of aspects of BLAST's output
when it is used to map Solexa Next-Gen DNA sequences 
to the human genome.

\noindent
{\bf Keywords:}
1000 genomes project 1KG,
genetic \mbox{programming}, GIP, GISMOE,
automatic software \mbox{re-engineering},
SBSE,
software modelling,
collaborative coevolution,
teams of agents,
soft memory, 
bioinformatics,
approximate string lookup,
sequence alignment,
SNP
\end{abstract}


\section{Introduction}
\label{intro}

Hand made software is expensive.
Therefore to make the widest possible use of it
we usually try to make programs as generic as possible.
Even though a task specific program might be better at that task,
it is often too expensive to write bespoke software 
and the user must make do with the generic program.
However if automated software production were to be successful,
the balance could turn from single generic programs to multiple
bespoke programs, each tailored to a particular task
or even tailored to each user.

We study a popular generic program BLAST \cite{ncbi_blast} 
to investigate the scope for
automatically customising it for a popular current demand,
mapping next generation DNA sequences.
BLAST is too generic, too slow to be used with Next-Gen sequences.
Indeed they were unknown when it was first written.
Yet matching them against the human genome is a vital component of an
increasingly important part of modern biology and 
doubtless this will also shortly play a vital role in medicine.
If it was practical,
BLAST would be the tool of choice for Next-Gen sequence alignments.
However it is far too slow to cope with the hundreds of millions of
DNA sequences generated by current,
let alone future,
sequencing machines.
Instead new hand made programs are used.
The new software re-uses ideas from BLAST
(e.g.\ string hashing and partial matches) 
but sometimes also includes newer ideas
(e.g.\ data compression algorithms
\cite{Langmead:2012:nmeth,Langdon:2013:ieeeTEC,Langdon:2013:GECCOlb}).
Also often the new performance comes at the price of losing some
functionality
\cite{Langmead:2009:GB}.
In the future
both combining existing code in new ways and
performance tradeoffs 
may be within the scope of software optimisation techniques.
In detail,
we show genetic programming (GP) can model aspects of BLAST's output
which are specific to Solexa Next-Gen DNA sequences 
and the human genome.

BLAST is perhaps the most important Bioinformatics tool
\cite[p848]{Bioinformatics-1998-Karplus-846-56}%
\footnote{%
Google Scholar says
BLAST 
had been cited more than 47\,000 times
and it is top of the
Nucleic Acids Research journal's (1~June 2013)
list of most cited articles.}.
Originally designed to find optimal matches between protein and amino
acid sequences, it and its variants are now widely used for many kinds
of sequence comparison,
in particular for DNA sequences.
With the advent of the sequencing of the human genome
and next generation sequencing
the demands of sequence matching have changed enormously
since BLAST was written.
With this in mind we have been investigating to what extent it might
be possible to automatically re-engineer a bespoke version for BLAST
for Next-Gen DNA sequence look up.
We are somewhat short of our goal of having evolutionary computing
rebuild a version BLAST which is more efficient at
aligning next generation DNA sequence with the human genome.
However there are certain aspects of the input data which are critical
to BLAST's performance which we have been able to model with 
genetic programming.
These include:
the quality and length of the match found
and indeed the number of matches found
or even if the DNA sequence does not occur in the human genome.

In the next section we briefly summarise 
BLAST and related programs, 
including important properties
that influence run time.
Then in Section~\ref{sec:data} we will describe 
the source of the training and verification data.
(All of these data are available online,
e.g.\
via
ftp.\allowbreak 1000\allowbreak genomes.\allowbreak ebi.\allowbreak ac.\allowbreak uk.)
Section~\ref{sec:results} shows it is possible 
(using genetic programming, Section~\ref{sec:gp})
to estimate in advance important properties of the BLAST's output.
Although GP has been used
many times before
this is the first time it has been used to model
a substantial man made software tool like BLAST%
\footnote{%
BLAST contains about 20\,000 lines of C++.}
as part of plan to reverse engineer it.

\section{Background}

In June 2013 
\href{http://en.wikipedia.org/wiki/List_of_sequence_alignment_software}
{Wikipedia}
listed more than 140 Bioinformatics tools which perform some
aspect of string analysis either on protein databases or
DNA sequences.
This represents a huge investment in manual labour.
Many of these tools can be regarded as finding a 
specific trade-off between performance
(typically in terms of accuracy and number of matches found)
and non-functional requirements
(typically speed, or speed and memory requirements).
As part of the
\href{http://www.cs.ucl.ac.uk/staff/W.Langdon/gismo/}{GISMOE}
program \cite{Langdon:2012:mendel,Harman:2012:ASE}
\mbox{\cite{Harman:2013:STTT,%
Petke:2013:SSBSE,%
langdon:2010:cigpu,Langdon:2013:GECCOlb,Langdon:2013:ieeeTEC}}\linebreak[4]
to mechanise aspects of software production we
are investigating to what extent
these trade-offs can be automated.

The performance of many sequencing tools is dominated by the length
and number of exact matches.
Run time can grow rapidly, e.g.~O($n^{3}$),
with number of matches~($n$)
\cite{Langdon:2013:ieeeTEC}.
Hash techniques allow exact matches to be found very quickly
but approximate matches are time consuming.
Therefore we investigated if GP could say in advance if a look up
is likely to match exactly and with what length of match
and if multiple matches would be encountered.
In sequence alignment multiple matches are less useful 
but are very time consuming,
thus (since they have redundant data)
biologists often ignore such sequences
\cite[page~1]{Cheung:2011:NAR}.

Where redundant data are available,
if we could identify high quality matches 
(cf.\ E~value, Section~\ref{sec:gp:training})
in advance
it would make sense to use them first.
Also
if we suspect a sequence will not match
it can be given a lower priority.
Similarly if it occurs more than once in the human genome
the problem of deciding how to resolve this ambiguity can be delayed
until after more promising sequences have been tried.
Finally 
perfect match ``seeds'' 
are
critical components of sequence look up.
Knowing in advance 
how long an exactly matching region
is likely to be
would make hash based search more efficient.

Although we model BLAST,
since it incorporates many Biological sequencing heuristics
and is widely used,
it has become the de~facto standard for sequence matching which
other tools seek to emulate.
Thus a model of BLAST is also a model of the desired output of 
other (more modern) sequence look up tools.
However although GP models are fast enough to make prioritisation
of DNA sequence matching feasible,
BLAST itself is still not competitive with more modern tools 
specifically designed for Next-Gen sequences.

\section{Training Data -- NCBI Human Genome, BLAST and 
the 1000 Genomes Project}
\label{sec:data}

\noindent
We used the official USA's National Center for Biotechnology Information (NCBI)
release of the reference human genome.
(
Release~37 patch~5, GRCh37.p5,
was down loaded
via \href{http://www.ncbi.nlm.nih.gov/blast/docs/update_blastdb.pl}
{update\_blastdb.pl} 
from http://www.\allowbreak{}ncbi.\allowbreak{}nlm.\allowbreak{}nih.gov/blast/docs.)
Similarly we used the official NCBI 
64~bit Linux version of BLAST (version 2.2.25+
was downloaded 
via ftp://ftp.ncbi.nlm.nih.gov/\allowbreak{}blast/executables/blast+/)

The 1000 genomes project \cite{nature09534}
is a consortium in which DNA from more
than one thousand individuals has been sequenced by 
several institutions using a variety of scanners.
To avoid unknown data inconsistencies during training
we decided to concentrate upon a
single scanner used by one laboratory.
Similarly we wanted to minimise true biological variation
and concentrated upon the software technology.
Therefore we chose one well studied family
and the Solexa data provided by the Broad Institute,
Cambridge, MA\@. 
This gives a large pool of homogeneous data.
The Solexa data includes sequences of a number of lengths
(from 25 to 489 DNA bases)
and so we further limited ourselves to 
all the Solexa data with exactly 36 DNA bases per sequence.
These sequences are ``single ended'' and give ``high coverage''.
We initially trained our GP using data from 11 different \mbox{Solexa-3623} scans
of the same female (NA12878) from Utah
(a total of
89\,087\,344 DNA sequences%
).

The Solexa data quality is highly variable.
In addition to its quality indicators,
the Solexa data use ``N'' to indicate a DNA base which it cannot
decide which of the four bases (A, C, G, T) it really is.
In one dataset less than 1 in a thousand sequences has an N.
In the worst training dataset (SRR001752),  
every record had at least one (typically two or three).
We selected data from the best,
the worst and an intermediate dataset for training.

\pagebreak[4]
\section{Genetic Programming}
\label{sec:gp}

This section describes in sufficient detail to allow reproduction of
our results the genetic programming system we used.
Section~\ref{sec:results} (page~\pageref{sec:results})
describes the models it evolved,
their performance
and also explores how they work.

\subsection{Preparing the Training Data}
\label{sec:gp:training}

\label{p.XY}
\noindent
Solexa scanners optically 
read DNA sequences from 60 or more square tiles.
It has been suggested that data quality may be worse near the borders
of the tiles.
However we do not see this effect. 
Nevertheless
for each of the three 
training datasets, 
we selected sequences in narrow (11 pixels wide) vertical 
and horizontal (10 pixels) strips 
crossing the whole of a tile (at pixel 511,779)
(Tile 119 was randomly selected for two scanner runs
but proved to give a small data sample in the third case,
so tile 78 was randomly chosen for the last set of NA12878 training data.)
This gave 
606,
565 and
1186
(total 2357)
DNA sequences each with 36 bases for training.

In addition to BLAST's default parameters,
we used  
{\tt -task blastn-short}
and {\tt -num\_threads~6}, 
as it looked up each DNA sequence
in the NCBI
reference human genome.
(Using 6 threads BLAST took approximately 7~hours
on a 32GB 8~CPU 3GHz Intel server
to process the 2357 sequences.
The GP runs used a single CPU each.)
In total for the 2357 queries
BLAST generated 14.5 gigabytes of output
containing 
60\,151\,902
partial matches.
For each partial match BLAST supplies a number of statistics.
These include its expectation (or E-value),
which is an estimate of how likely the match would be at random.
(The E-values lie in the range $10^{-10}$ to 6.6.)
BLAST also reports 
the number of bases which match exactly between the query sequence and
the human reference genome (0-36).

The NCBI reference database for the human genome contains DNA
sequences from a number of sources
\linebreak[4]
\cite{Astarloa:2009:BT,langdon:evobio12}.
In many cases these sources overlap.
Therefore to count the number of matches found by BLAST,
we include only high quality matches 
(i.e.~\mbox{E$<\!10^{-5}$})
and restrict BLAST to the
reference sequence for
human chromosomes~1 to~22 and
the human X and Y chromosomes.

\subsection{Randomised Test Suite Sub-sampling}
\label{sec:RndChoices}

\noindent
When a large volume of training data is available
we had previously used a random sub-sample 
of the test data at each generation 
\cite{langdon:2010:eurogp}
to reduce the volume of testing but also
found it helped with generalisation.
Here we also use it to spread the training data more uniformly.

We divided the training data 
into non-overlapping bins
using the value to be predicted.
(In the case of the two binary classification problems,
Sections~\ref{sec:high_quality} and~\ref{sec:N4},
there are just two bins.)
Each generation
equal numbers of training examples are randomly chosen from each bin.
Where a bin contains more examples than are needed 
the examples are kept in the same random order and the next group
are taken.
Except where noted,
this ensures the examples used in the next generation are not the same
ones as used in the previous generation.
If there are insufficient examples left,
the bin's examples are first put in a new random order.
(This is somewhat reminiscent of Gathercole's 
DSS \cite{gathercole:1994:stss},
as used commercially 
\cite{foster:2001:discipulus}.)

\subsection{Predicting the Quality (E~value) of DNA Matches}
\noindent
Typically BLAST reports many matches.
However it
sorts them 
so that the best match
comes first.
Therefore we simply took the E~value of the first match
to be GP's target value.
Where BLAST was unable to find any match
this was presented to GP as an E value of 100.
Due to the wide range of E values,
GP works with $\log_{10}$(E) rather than the E value directly.
Using the integer part of $\log_{10}$(E)
we divided the training data into 13 groups
(separated by powers of ten).
Each generation 35 examples are randomly chosen from each group.
This means each generation $12\times 35+11 = 431$ (of 2357) examples are used.
(The twelfth bin contains only 11 examples.)
Note,
the bins are only used for training,
data used to test the evolved predictors are drawn uniformly.
The GP fitness function is the Pearson correlation between the value
calculated by GP and $\log_{10}$(E).

\subsection{Predicting the size of the Best BLAST Match}
\noindent
Again we took the best match
to be GP's target
and this time asked GP to predict its length
(i.e.~the length of the first match
reported by BLAST)\@.
There are 137 Solexa training sequences where BLAST was unable to find any
match. 
These are treated as having a length of zero and placed in a group by themselves.
All reported matches have a length of at least 18 bases.
They are placed in ten groups: 
18-19, 20-21, 22-23, ..., 34-35 and length 36.)
So $11\times 35=385$ 
of 2357 examples are randomly chosen for use in each generation.
Fitness is the correlation between GP's value
and the actual length.

\begin{table*} [tbp]
\setlength{\temp}{\textwidth}
\settowidth{\tempa}{Result Terminals:}
\addtolength{\temp}{-\tempa}
\addtolength{\temp}{-1\tabcolsep}
\settowidth{\tempa}{M Sum0}
\caption{\label{gp.details}
GP Parameters 
for Predicting BLAST results with 1000 Genomes Project
Solexa short DNA sequences.
}
\begin{tabular}{@{}l@{ }p{\temp}@{}} \hline
Pass0 Terminals: 
& 
1037 random constants
pos len=36 A C G T Self Complement Samesize Opposite
N Qual 
S Run CountN
X Y
Aux1 Aux2
\rule{\tempa}{0pt}
\\
Pass1 Terminals: &
1037 random constants
pos len=36 A C G T Self Complement Samesize Opposite
N Qual 
S Run CountN
X Y
Aux1 Aux2
M 
Sum0
\\
Result Terminals: &
1037 random constants
len=36
S Run CountN
X Y
Aux1 Aux2
Sum0 Sum1
\\
Functions: & IFLTE
ORN ADD SUB MUL DIV
LOOK$_{01}$
set\_Aux1 set\_Aux2 
sum\_Aux1 sum\_Aux2 
\\
Fitness:       & If predicting E or length, fitness is correlation.
If classifying high quality match or repeated high quality match then
fitness is the
number of correct answers
\\ 
Population:    & Panmictic, generational. 10\,000 members.
4 members per selection tournament.
New training sample each generation.
\\ 
Parameters:    & 
Initial pop ramped half-and-half 6:2.
50\% subtree crossover, 
22.5\% point mutation,   
22.5\% mutation which swaps constants,  
2.5\% shrink mutation,   
2.5\% subtree mutation. 
No depth limit,
max size 1022.
Stop after 100 generations.
\rule[-6pt]{0pt}{6pt}
\\ \hline
\end{tabular}
\end{table*}

\subsection{Predicting High Quality BLAST Matches}
\label{sec:high_quality}
\noindent
We evolve two classifiers for the number of BLAST matches.
The first (this section) returns a positive value if GP thinks
at least one high quality match will be found
(i.e. \mbox{${\rm E}<10^{-5}$}).
Conversely,
there are 
997
(42\%) 
training DNA sequences which either BLAST could not find any
matches
or where the matches had poor E values (E$\ge 10^{-5}$).

With both classifiers
(i.e.~this section and the next),
each generation 300 positive and 300 negative 
randomly chosen examples are used to assess every 
individual's fitness.
Fitness is the number of training cases 
correctly classified.

\subsection{Predicting Repeats in the Human Genome}
\label{sec:N4}

When predicting repeats,
a positive value is used to indicate GP 
expects there to be multiple instances
of the DNA sequences in the Human genome.
(Repeated sequences are quite common 
\cite{ihgsc:2001:nature}.)
However
there are only
324 training
sequences with multiple high quality 
matches.
Selecting 300 positive examples
means almost all of them are used every generation.
In contrast each of the 2033 negative examples
will have to wait at least six generations 
($\lfloor 2033/300 \rfloor=6$) 
before it is even eligible to be reused by the fitness function.
(On average each negative example is used in 16 generations.)
As with all our other predictors,
the separate data used to validate the evolved predictor are drawn uniformly.

\subsection{Genetic Programming Architecture and Primitives}

The rest of Section~\ref{sec:gp} describes the GP
representation, data flows within its three components,
terminal and function sets
and the speed of the GP system.
We used a multiple tree GP (based on \cite{langdon:book})
with functions and terminals inspired by Koza's transmembrane
prediction work \cite{koza:gp2}.
(Details are given in Table~\ref{gp.details}.)

\subsubsection{Co-Evolving Three Trees}

\begin{figure*} 
\centerline{\includegraphics{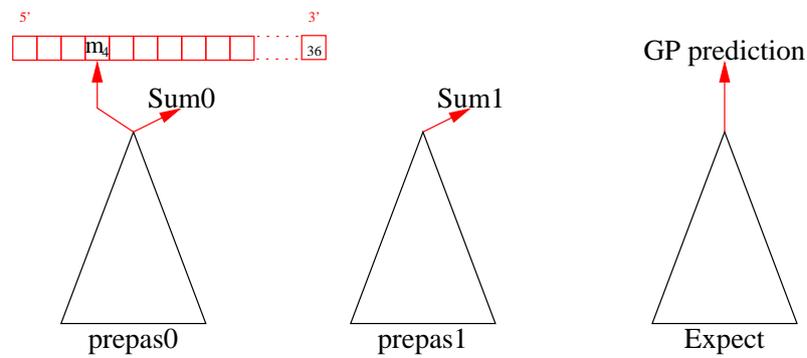}} 
\caption{\label{fig:3trees}
Schematic of three tree architecture used to evolve prediction of 
BLAST results.
Tree prepas0 is called once 
for each base in the DNA sequence in 5'--3' order.
Various memory cells (not all shown)
allow it to communicate partial results both
with itself and with the other trees.
Similarly the evolved tree prepas1 is called once per base and
can access information about the DNA sequence and results passed to it
by prepas0.
Finally the Expect tree is called once. It can use data provided by 
both prepas0 and prepas1 to return the overall GP prediction for the
whole DNA sequence.
Figures 
\protect\ref{fig:SRR00three.4.outE4_2b_gp.load},
\protect\ref{fig:SRR00three.4.outL4b_gp.load},
\protect\ref{fig:SRR00three.5.outZ4b_gp.load}
and~\protect\ref{fig:SRR00three.1.outZ4b.100000.approx}
contain examples of actual evolved code.
}
\end{figure*}

\noindent
It seems clear that the evolved classifiers will need to scan the DNA
sequence.
It is unclear how many times they will want to scan it.
To support serial processing of the sequence and 
still allow evolution some freedom of choice,
we adopted a novel three tree architecture
(Figure~\ref{fig:3trees}).
The first two trees scan the sequence and pass data via memory to the
final one.
GP is free to decide how to use this architecture.
It can readily use it to process the complete DNA sequence 
0, 1 or 2 times.
The first tree is called once for each member of the sequence 
from the start to the end.
(In Bioinformatics this is known as 5'--3' order).
Thus the first GP tree is used 36 times.
Then the second tree is called for each DNA base in the sequence in the
same order.
(Again making 36 times per fitness evaluation.)
Finally the last tree is called once and its result is used to
determine the fitness of the whole team.

The trees can also be thought of as team members or agents.
They are independently evolved in that crossover and mutation
act on a single agent but
whole teams are forced to coevolve collaboratively.
Each member is locked into its team and cannot act outside it.
In earlier experiments with evolving memory 
we showed GP can coevolve such teams of agents.
\cite{langdon:book} contains examples of teams of five and 15 agents.
Recently GP has since been used to evolve even bigger teams
\cite{conf/iwcls/Moreno-TorresLGB09}.
GP can evolve individuals which pass 
results returned by 
earlier tree calls and previously calculated intermediate values
amongst team members by
using a variety of special purpose and general purpose memory cells.
(As is common in machine learning,
GP individuals cannot affect the fitness of other members of the
population or influence their own behaviour on other test cases.)

\subsubsection{Passing Data Between Trees via Memory}
\label{sec:memory}

\noindent
It is expected that passing along the DNA sequence will enable GP
to calculate complex statistics somewhat similar to running averages.
To assist this the sum of the 36 results produced by each of the
prepass trees is stored and made available to the following parts of
the GP individual via read-only terminals Sum0 and Sum1.

The GP leaf M refers to 36 read-only memory cells,
which allows the second prepass to use the result returned by the
first pass.
It can be thought of a similar to an ADF \cite{koza:gp2}
with a cache \cite{langdon:book}.

We also have more conventional indexed memory
\cite{kinnear:teller}.
There are two floating point memory cells,
called Aux1 and Aux2,
and routines to write to them, set\_Aux1 and set\_Aux2.
As with \cite{langdon:book}, 
functions set\_Aux1 and set\_Aux2
take one argument, which is assigned to the memory cell
and return the memory's new value.

Inspired by the work on ``memory with memory'' 
\cite{Poli:2009:JAEA}
we also introduce two functions:
\linebreak[4]
sum\_Aux1 and sum\_Aux2.
Instead of overwriting their memory cell's contents they add
their input's value to it.
(Again returning the memory's new value.)
Alternatively these can be thought of as parametrised
memory increment instructions
\cite{langdon:book}.

\subsubsection{DNA bases}
\label{sec:bases}

\noindent
Inspired by Koza's work on protein sequences \cite{koza:gp2}
we use five terminals
(A, C, G, T and~N)
to indicate (in both prepass trees)
if the current bases is an A, C, G or~T\@.
N~indicates the scanner did not know which of the four bases it
is.
The leaf has the value 1 if is true and \mbox{-1} otherwise.
(Thus leaf N is always -1 unless the DNA scanner has indicated it
cannot tell which of the four bases it is.)

\subsubsection{Look Ahead}
\noindent
It may be convenient for GP to be able to compare nearby parts of the
sequence.
The function LOOK moves the active position (denoted by terminal pos)
ahead one.
Within the overall limit on program size,
LOOK can be as deeply nested as evolution likes
and so it allows arbitrary look ahead.
I.e.\ special inputs, like A, when invoked by LOOK's single argument,
respond as they would do when the prepass tree is called later along
the sequence.
LOOK applies to all position dependent primitives
(including Qual, Run, M, S and pos itself).
Protection against looking pass the end of the DNA sequence is given.

\subsubsection{DNA Specific Terminals}
\noindent
The complementary binding of A--T and C--G is well known.
We sought to give ready access to such information.
The four terminals:
Self, Complement, Samesize and Opposite
each compare the current base (as updated by LOOK)
with the base at the standard position (i.e.\ without LOOK).
Like A, C, G, T and N, they signal true and false with 1 and -1.

Self returns 1 if the current DNA base and the 
``LOOKed'' at DNA base
are the same.
Complement returns~1 if they are members of a complementary pair.
Samesize returns 1 if they contain the same number of rings.
(C and T are ``small'' and have one ring.
A and G are large and have 2 rings.
Arbitrarily N is given 0 rings.)
Opposite is true only if Self, Complement and Samesize are all false.

\subsubsection{Solexa Quality Terminals}
\noindent
As well as N, the Solexa scanner includes a quality
value for each base. 
These 
are presented to the GP via the Qual terminal as
values in the range 0, 0.1, 0.2, ..., 3.9, 4.0.
(4.0~indicates the DNA scanner has the highest confidence in its
output.
0~means it has no confidence in it.)

\subsubsection{Runs of the Same DNA Base}
\noindent
It is known that sequences of the same base can affect Bioinformatics
equipment \cite{upton:2008:NGT},
so we provide GP with a primitive which counts
the length of identical bases up to and including the current point.
(Run's value is therefore in the range 1, 2, 3, ... theoretically
up to 36.)
No special treatment for N is provided.
I.e.\ an N value terminates a run of other letters 
but we can also have runs of Ns.

\subsubsection{CountN}
CountN
is simply the number of ``N'' 
(i.e.\ unknown)
bases in the current DNA sequence.
Typically CountN takes the value 0, 1, 2, or~3.
Since CG ratio is known to be important 
to the Solexa scanner
\cite{Cheung:2011:NAR},
we might also have provided similar 
CountA, CountC, CountG and CountT terminals.

\subsubsection{Entropy S}
\label{sec:S}

\noindent
The terminal
S holds the entropy or information content of the string
in bits 
from its start up to the current position.
I.e.\
S$=\sum_{A,C,G,T} - p_i \log_{2}(p_i)$.
Where, for example, 
$p_A$ is the number of As in the current DNA sequence from its start
up to the current position 
divided by the number of all four bases up to the current position.
Since N indicates an unknown base, it is accounted for
by adding
$1/4$
to each of the four bases' totals.
S~in the third tree yields the entropy of the whole sequence.

\subsubsection{X and Y}

\noindent
The leafs X and Y tell the GP where the DNA sequence was located on
the Solexa tile.
(See Section~\ref{p.XY}).
They are both normalised to the range 0..1
by dividing by the width of the tile in pixels.

\subsubsection{Ephemeral Random Constants}
\noindent
1000 values were randomly selected from a tangent distribution
\cite{langdon:book}.
(I.e.\
a value is randomly chosen from the range $0..\pi$
and its tangent is taken.)
This gives a few very large numbers 
(the largest was 632.124324
and the smallest -425.715953)
but about half the values lie in the range $-1..+1$.
This was supplemented by the 37 integer values: 0, 1, ... 36.

\subsubsection{Functions}
\label{sec:funcs}
\noindent
In addition to LOOK and the four usual arithmetic
operations 
(+, -, $\times$ and protected division, which returns 1 on
divide by zero)
we include if-less-than-or-equal
\cite{langdon:book}
and Koza's ORN \cite{koza:gp2}.
ORN takes two arguments.
If the first is true (i.e.~$>0$)
the second is skipped and  ORN returns 1.
Otherwise it evaluates its second argument and returns 1 if it is
true.
If the second argument is false (i.e.~$\le 0$),
ORN returns~-1.

Mostly default behaviour is applied for special values like NaN or
infinities.
However care is needed with calculating correlation based fitness
and protecting against rounding errors etc.\
causing numerical instabilities 
(e.g.~%
leading to negative variance).
Members of the GP
population causing unresolved numerical problems are given
low fitness values.

\subsubsection{GP Speed}

\noindent
The special GPquick interpreter 
\cite{singleton:byte}, 
\cite{langdon:book}
we used
processes on average 213 million GP primitives
per second.
In \cite{langdon:2010:eurogp} we reported a GPU based GP interpreter
running more than a thousand times faster,
however this was build to exploit 
both the graphics hardware accelerator and the
bit level parallelism inherent in
Boolean problems \cite{poli:1999:aigp3},
rather than floating point numbers as used here.
Our GPquick compares well with tinyGP~\cite{poli08:fieldguide},
which processes in the region of 
20--80 MGPops$^{-1}$ 
(C~version). 
\cite{langdon:2011:SC} discusses the speed of recent compiled and
interpreted GP approaches, particularly those using graphics
hardware accelerators.
(See also \cite[Table~15.3]{langdon:2013:ecgpu}.)

\pagebreak[4]
\section{Evolved Prediction of BLAST Matches}
\label{sec:results}

\begin{figure} 
\centerline{\includegraphics{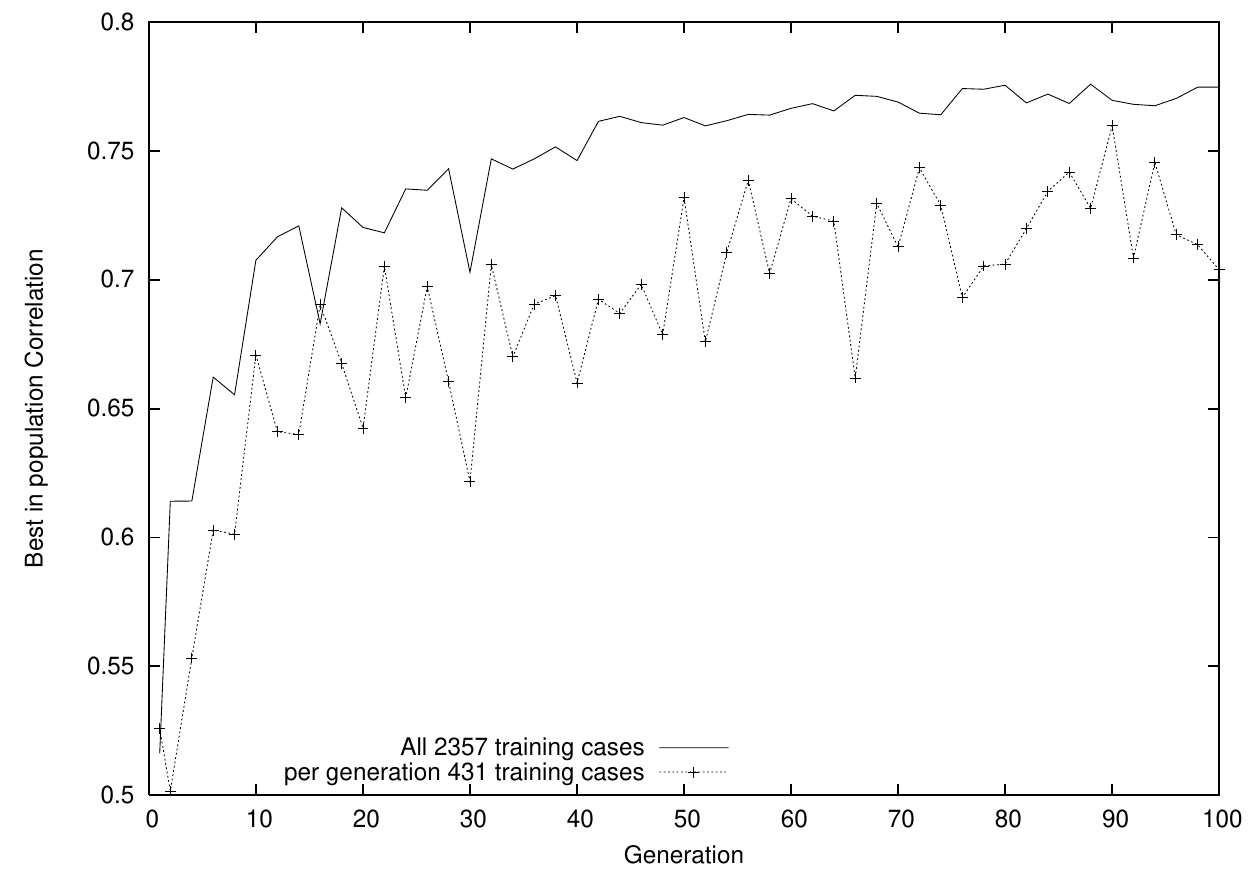}}
\vspace{-2ex}
\caption{\label{fig:expect}
Evolution of performance on training data
of best of generation BLAST E-value predictor. 
Overall training performance (lines) follows
performance on 431 tests being used in the current generation
(lines with $+$).
}
\end{figure}

\noindent
In each of the four problems,
we extracted the best of generation 
individuals at generation ten and generation one hundred.
(Details are given in Table~\ref{gp.details}.)
Figure~\ref{fig:expect} shows the evolution of the best of
generations fitness, 
both on its per generation training data (lines with~$+$)
and on the whole of the training data
(solid lines).

In each of the four prediction tasks and at generation ten 
and generation one hundred,
we tested the smallest of the 
best of population individuals
on a validation set of 31\,000 Solexa  DNA sequences.
The validation data came from completely new sets of 
sequences firstly chosen uniformly at random
from the eleven 
NA12878 data sets (a~1000 from each).
There were only two other individuals 
in the 1000 genomes project
with more than ten data scans
from similar Illumina2 Solexa scanners.
These are NA12891 and NA12892,
who are her father and mother respectively.
Ten data scan were chosen at random for each and then
1000 DNA sequences were chosen from each scan.
(This entailed running BLAST in total another 31\,000 times,
taking about 108 hours.)

\begin{figure*} 
\centerline{\includegraphics{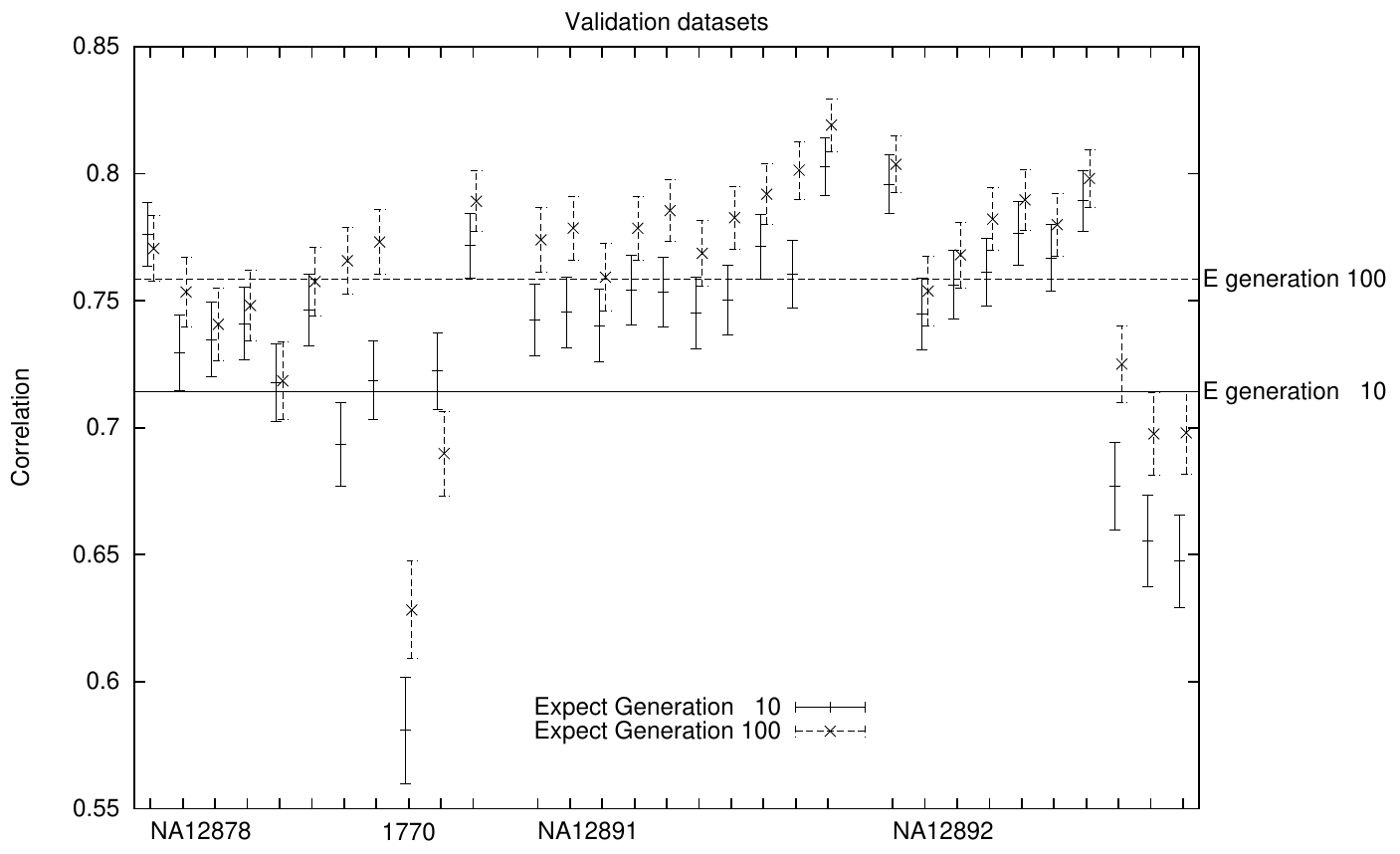}}
\caption{\label{fig:verify_corE}
Validation of evolved predictors of
BLAST E-values 
evolved at generations~10 and~100
on 31 DNA scans from three individuals (NA12878, NA12891, NA12892).
Apart from NA12878's DNA scan SRR001770, 
performance
on unseen data
is close to that on 2357 training data
(horizontal lines).
Error bars indicate standard error \protect\cite[p212]{hotelling:1953:JRSSb}.
For comparison the
horizontal lines indicate performance on the 2357 training
data.
}
\end{figure*}

\begin{figure*} 
\centerline{\includegraphics{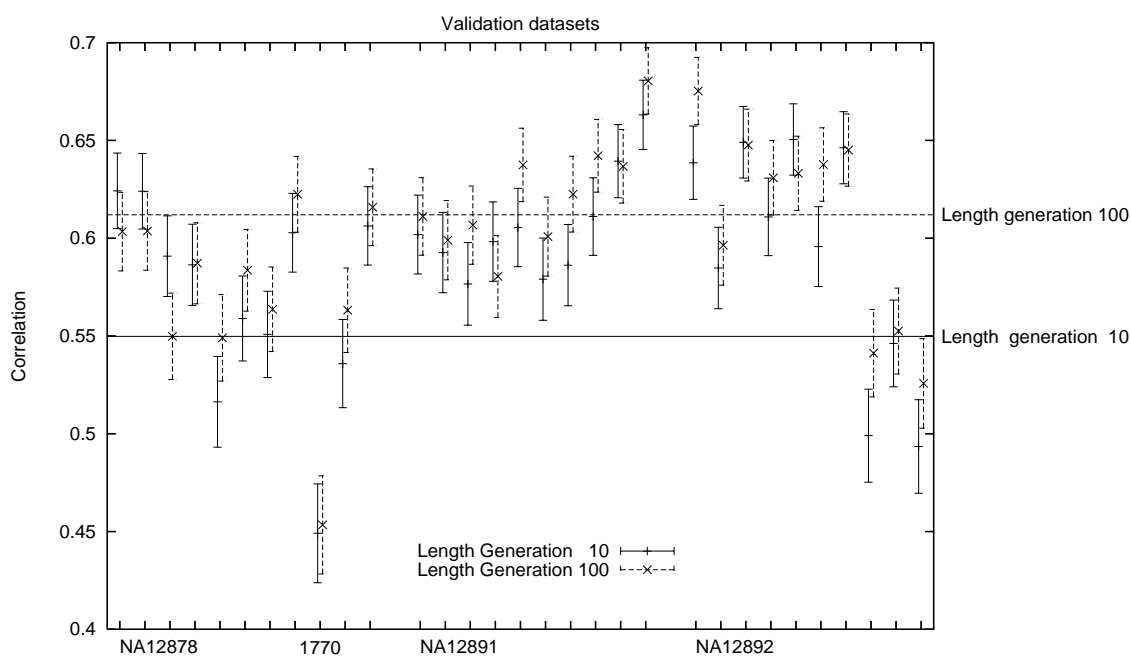}}
\caption{\label{fig:verify_corL}
Validation of evolved predictors of
length of BLAST matches
from generations~10 and~100
on 31 DNA scans from three individuals (NA12878, NA12891, NA12892).
As with predicting E~values,
Figure~\protect\ref{fig:verify_corE},
apart from NA12878's DNA scan SRR001770, 
performance
on unseen data
is close to that on 2357 training data
(horizontal lines).
Error bars are standard error.
}
\end{figure*}

\begin{figure*} 
\centerline{\includegraphics{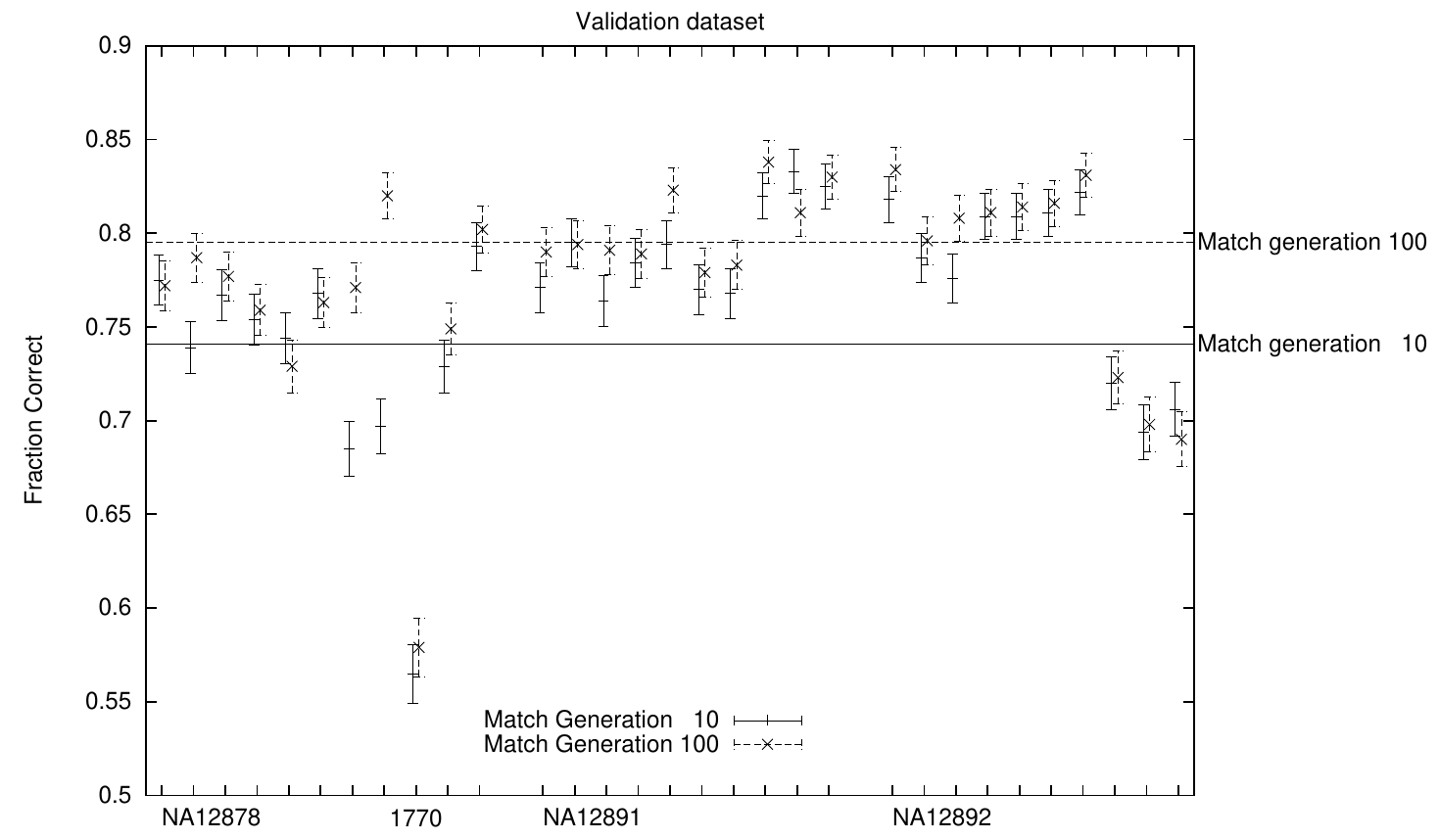}}
\caption{\label{fig:verify_perMa}
Performance when predicting whether BLAST will find matches or not
on unseen data.
Apart from NA12878's DNA scan SRR001770, 
performance is close to that on the 2357 training data
(horizontal lines).
As with Figures~\protect\ref{fig:verify_corE} 
and~\protect\ref{fig:verify_corL},
we plot validation scores for predictors
evolved at generation~10 and at generation~100.
Error bars are standard error.
}
\end{figure*}

\begin{figure*} 
\centerline{\includegraphics{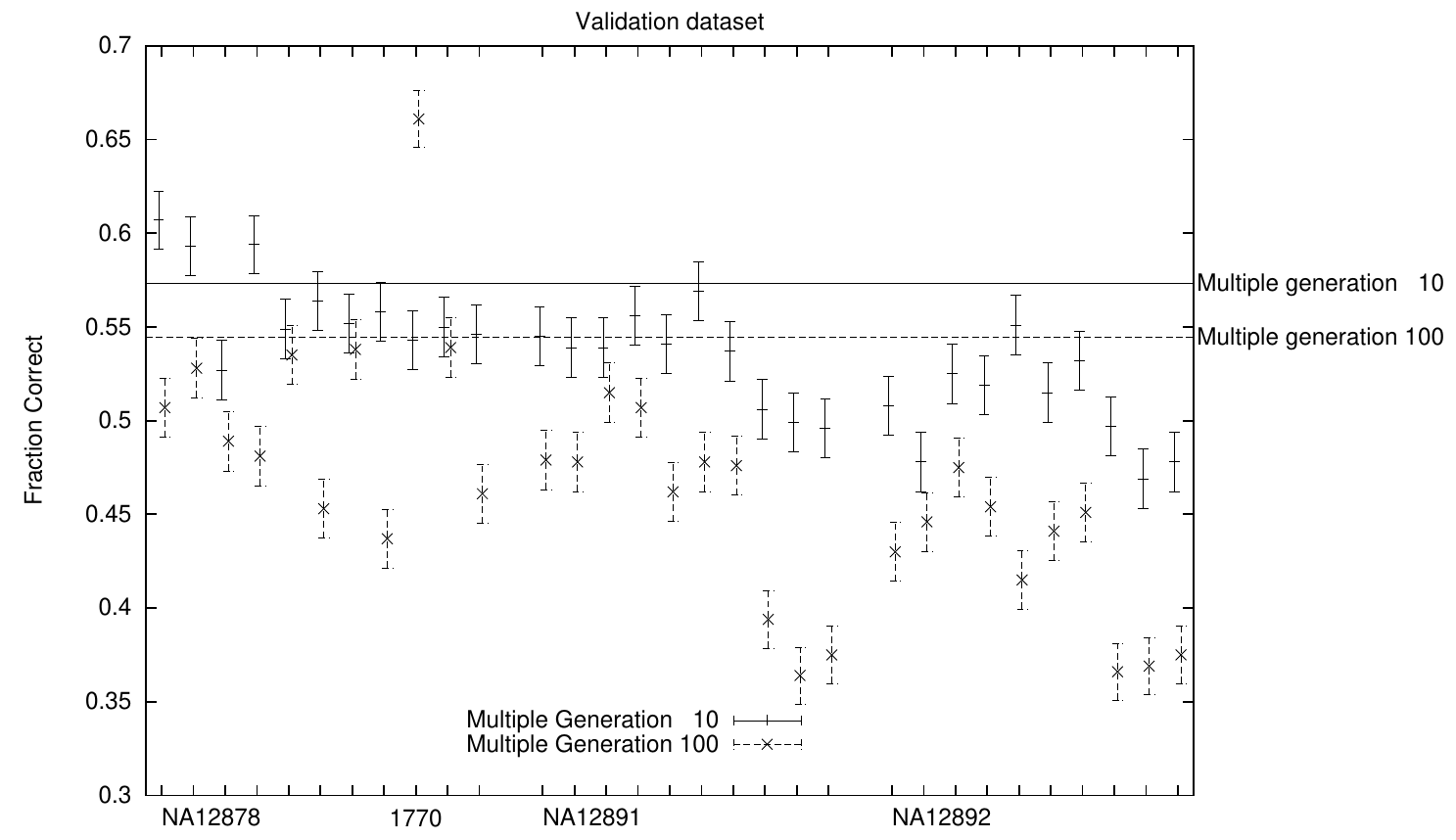}}
\caption{\label{fig:verify_perMu}
Predicting repeated matches in the validation data.
As with Figures~\protect\ref{fig:verify_corE},
\protect\ref{fig:verify_corL}
and~\protect\ref{fig:verify_perMa},
we plot validation of GP results,
as evolved at generations~10 and~100.
Unfortunately,
particularly for generation 100,
performance is worse on the validation data.
However
Figure~\protect\ref{fig:verify_multiple} shows
this is compensated for a little by GP's being able to predict
most of the minority class.
Again horizontal lines show performance on the 2357 training data.
Also NA12878's DNA scan SRR001770 is again not typical.
(Actually SRR001770 is better than average in this case.)
Error bars show standard error.
}
\end{figure*}

In all four cases
GP was 
to some degree successful
at predicting BLAST's output.
Although there are statistically significant differences between the 
thirty one validation DNA scans
(e.g.\ SRR001770),
performance of the evolved predictors held up well.
See Figures~\ref{fig:verify_corE}, \ref{fig:verify_corL},
\ref{fig:verify_perMa}, \ref{fig:verify_perMu}
and~\ref{fig:verify_multiple}.
In contrast to the other scans of NA12878's DNA,
scan SRR001770
has about half as many high quality matches.
(I.e.\
BLAST matches with
${\rm E} = 10^{-10}$.)
The reduction in high quality matches,
being approximately the same as the increased fraction of
intermediate quality matches,
i.e.~%
${\rm E} \approx 10^{-7}$.
Even on the most difficult problem,
predicting
the existence of repeated short sequences in the Human genome,
across all 31\,000 verification sequences,
GP's predictions are much better than random guessing,
\mbox{$\chi^2$ = 108} 
(1~dof).

\pagebreak[4]
\subsection{Predicting BLAST ``E'' Values}
\label{sec:E_results}
\noindent
After 
100 generations 
GP evolved a program 
whose correlation with BLAST's \mbox{E-value}
is r=0.76 
on the training data
and whose median correlation 
is r=0.77 
on all 31\,000 validation sequences,
see $\times$ in Figure~\ref{fig:verify_corE}.

Generation 100's
77 primitive program 
can be simplified to an almost equivalent program
(see Figure~\ref{fig:SRR00three.4.outE4_2b_gp.load})
\begin{figure*}
\footnotesize
\input{SRR00three.4.outE4_2b_gp.load}
\caption{\label{fig:SRR00three.4.outE4_2b_gp.load}
Simplified best of generation 100 BLAST E-value predictor.
GP gives its prediction via the value returned by the {\tt Expect} tree.
{\tt Sum0} contains the sum 
of the 36 values returned by
tree {\tt prepas0} when it is called in turn for each of the 36 bases in 
the DNA sequence generated by the Solexa scanner.
The average correlation with BLAST's E-value is 0.76. 
Much of its predictive power comes from the Solexa quality indicators
(via the {\tt Qual} terminals).
See Section~\protect\ref{sec:E_results}.
}
\end{figure*}
whose final output depends upon the sum over all 36 position in the
Solexa DNA sequence of the value calculated by prepas0.
The use of sum\_Aux2 at the root of prepas0 
frequently occurs and gives more weight to the 3' end of the sequence.
GP uses nested LOOKs to compare nearby DNA bases and their reported
quality. 
In fact the best of generation 10 individual from the same run
uses only Solexa's quality indicator to predict BLAST's E-value.
Although performance is not as good,
it still obtains 
 r=0.68 
on the training data
and is almost equivalent to
$\sum_{5'}^{3'} i $Qual $\times$ $\sum_{5'}^{3'}i/$Qual.
(Remember division is protected, Section~\ref{sec:funcs}.)
The formula suggests
lower quality Solexa data contains more mistakes
which in turn leads to BLAST matches 
where more DNA bases do not align exactly and hence BLAST assigns them
poorer E-values.
Notice again the 3' end is weighted 36 times more important than the 5' start 
of the Solexa DNA sequence.
This 
suggests that GP has learnt that the
probability of errors accumulate as the scanner moves along the
DNA sequence from the 5' start.
Thus GP assigns more importance to data from the 3' end of the
sequence than from the 5' start of the sequence.

The evolved model (Figure~\ref{fig:SRR00three.4.outE4_2b_gp.load})
gives a quick way of predicting how good 
the match between the DNA sequence and the human genome is.
Thus redundant sequences which are predicted to give poor quality matches 
could be given reduced priority or even ignored 
in favour of better DNA sequences generated by the scanner.

\subsection{Predicting Match Length}
\label{sec:len_results}

\noindent
Predicting the length of the match found by BLAST proved
more difficult with the smallest best of generation 100
predictor having a correlation on all the training data of 
r=0.61. 
However again this holds up well on the validation data,
with a median correlation 
r=0.60,
see $\times$ in Figure~\ref{fig:verify_corL}. 
The error bars confirm 
the results are highly statistically significantly
different from the null hypothesis
(no correlation).

\pagebreak[4]
The best of generation 100 program 
is shown in Figure~\ref{fig:SRR00three.4.outL4b_gp.load}.
\begin{figure*}
\input{SRR00three.4.outL4b_gp.load}
\caption{\label{fig:SRR00three.4.outL4b_gp.load}
Smallest
best of generation 100 predictor of the 
length of the best match found by BLAST\@.
GP's prediction is given by the {\tt Expect} tree.
In this program, 
the {\tt Expect} tree's return value is 
largely determined by the second prepass along the DNA sequence
(via variables aux1 and aux2 and {\tt sum1}).
{\tt Sum1} $= \sum_{i=1}^{36}{\tt prepas1}$\protect\rule[-6pt]{0pt}{6pt}.
Its predictive power comes to a large extent from the quality
(via {\tt Qual} and {\tt CountN} leafs)
of the Solexa data.
See Section~\protect\ref{sec:len_results}.
Although position data {\tt X} and {\tt Y} are included they play a
minor role. 
}
\end{figure*}
The best program at generation ten
has a similar performance (r=0.60) 
and being smaller it is easier to simplify.
It is similar 
to a predictor which returns
$\sum_{5'}^{3'}$ if(Qual$_{i}>$1) $i$ else -$i$.
Notice that again the 
3'~end 
is weighted 36 times more important than the 5'~start.
This formula means 
high quality levels (i.e.~bigger than~1.0)
especially near the end of the DNA sequence,
suggest longer BLAST matches
(r=0.54).   
The evolved formula includes
ignoring low quality values on A~bases.
If we add just this,
we can get back almost all the performance
(r=0.59).  
Whereas only using Qual$_{i}>$1 by itself,
excluding the location weighting~$i$, 
still gives a correlation of
r=0.48. 

Again we see the quality of the sequence generated by the scanner
playing a dominant role. Poorer quality sequences tend to have
more errors leading to BLAST finding sequences in the human genome
which do not match exactly. A single mismatch at the end of the query
sequence reduces the length of the reported match marginally,
whereas if its in the middle it can half the length of the exactly
matching region.

Typically sequencing tools like BLAST assume a minimum length of
exactly matching sequences.
(If the query sequence is disrupted so 
the number of consecutive matching DNA bases is less than this,
then the sequence will not be found.)
Sub-sequences of the minimum length are used as hash keys into the
reference genome.
Longer keys tend to mean the hash tables are smaller 
and there are fewer hits which speeds the search.
Also both fewer hits and longer hash keys
tend to mean the subsequent fuzzy matching has less work to
do and so is faster.
BLAST relies on the user to set the correct hash key size.
However users typically use the supplied default,
which is fixed (for a given class of queries).
If GP could tell us 
in advance how long the exact matching region was going to be,
the optimum hash key size could be used.

\subsection{Predicting High Quality BLAST Matches}
\label{sec:match_results}

\noindent
The smallest of the best of generation 100 programs evolved to predict
if BLAST will find a high quality 
match 
(Figure~\ref{fig:SRR00three.5.outZ4b_gp.load}),
gets
80\% 
of the training data correct.
The $\times$ plot in Figure~\ref{fig:verify_perMa}
shows 
this holds up well on the verification sequences
(median 79\%). 
Again this is highly statistically significant.
(Across all 31\,000 verification sequences
\mbox{$\chi^2$ = 8400}, 
1~dof).

\begin{figure*}
\input{SRR00three.5.outZ4b_gp.load}
\vspace*{-2ex}
\caption{\label{fig:SRR00three.5.outZ4b_gp.load}
Smallest
best of generation 100 predictor of whether BLAST will find a good match.
Here the evolved classifier has not been simplified.
Figure~\protect\ref{fig:SRR00three.1.outZ4b.100000.approx}
shows an almost equivalent program.
}
\end{figure*}

\pagebreak[4]
Since the generation 100 code is large
(Figure~\ref{fig:SRR00three.5.outZ4b_gp.load}),
we again start from the smallest best of generation~10 program
which is slightly less accurate
(74\%)
but smaller.
It can be simplified to
the code shown in 
Figure~\ref{fig:SRR00three.1.outZ4b.100000.approx}.
\begin{figure*}
\input{SRR00three.1.outZ4b.100000.approx}
\vspace*{-2ex}
\caption{\label{fig:SRR00three.1.outZ4b.100000.approx}
Simplified best of generation 10 predictor 
of whether BLAST will find a good match
(cf.~%
Figure~\protect\ref{fig:SRR00three.5.outZ4b_gp.load}).
Section~\protect\ref{sec:match_results} describes how it works.
}
\end{figure*}
The simplified code also has a training accuracy of 74\%.
Remember the prediction is given by the sign of the ``Expect'' tree
and so essentially the predictor works by comparing 
the sum calculated by the second scan of the sequence
with 72.0.
The second prepass calculates
\mbox{Sum1$=\sum_{5'}^{3'} (Qual_{i+1} - G_{i+1})/Qual_i$}.
Sum1 takes values spread widely around a median value of
71.1. 
The average value of S is 1.9 
so on average the last term 
\mbox{\tt (MUL S 1.177709)}
comes to 
about 2.2. 
This enables entropy (Section~\ref{sec:S})
to play a marginal role 
where $\sum_{5'}^{3'} (Qual_{i+1} - G_{i+1})/Qual_i$ is near its
average value
but the evolved predictor's performance comes mostly
from the prepass through the whole DNA sequence.
The prepass uses
the Solexa quality values and whether or not the next DNA base
is a G\@.
However 
the presence or absence of G bases is relatively unimportant.

Again we see the Solexa quality data playing a dominant role in
the evolved predictor.
Since, in this case, we are asking GP to predict the quality of the
match found by BLAST this is perhaps less surprising.
However evolution has managed to find other information and weave it
into the classifier to improve it above simply using the
quality indicators provided by Solexa.

\subsection{Predicting Repeats in the Human Genome}
\label{sec:repeats}

\noindent
After one hundred generations 
GP evolved a program which scores
54\% 
on all the training data.
Its median score is 
46\% on the validation data
(see $\times$ in Figure~\ref{fig:verify_perMu}). 
Although obviously reduced from its performance on its training data,
again this is highly significant.
Across all 
31\,000 verification sequences
\mbox{$\chi^2$ = 910} 
(1~dof).

\begin{figure} 
\centerline{\includegraphics{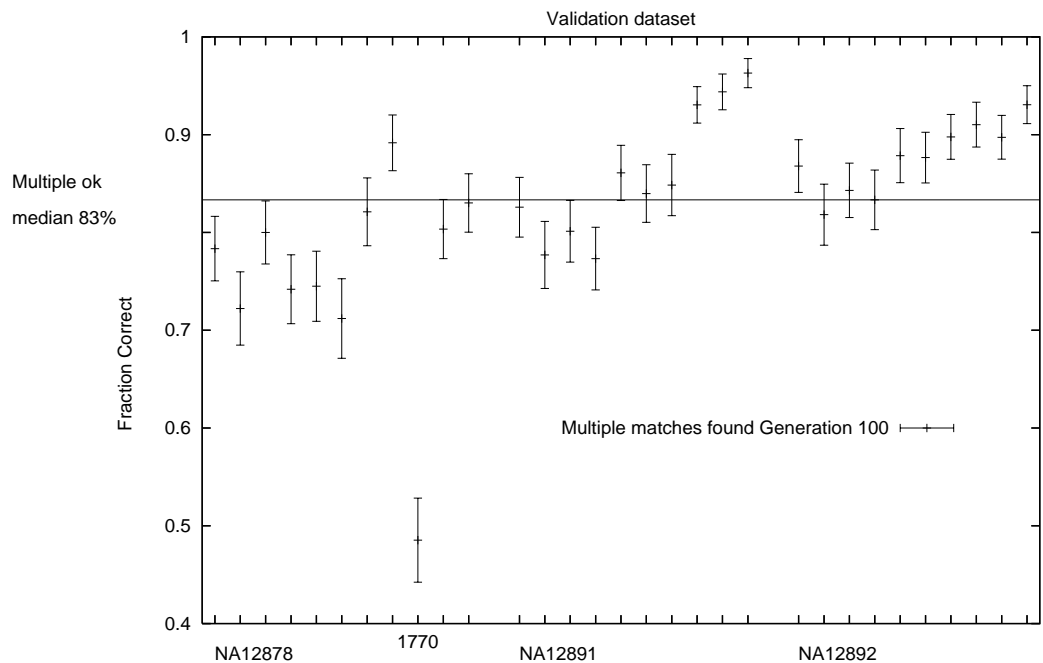}}
\caption{\label{fig:verify_multiple}
Plot on 31 verification datasets
showing GP
on average correctly predicts 
83\% of repeated  DNA sequences (the minority class).
Error bars indicate standard error \protect\cite[p212]{hotelling:1953:JRSSb}.
}
\end{figure}

Only
14\% of the training data 
give rise to multiple high quality BLAST 
matches.
So although each generation the fitness function tests the evolving
population on a 50:50 balanced dataset
(see Section~\ref{sec:N4})
good multiple matches are very much the minority class.
Despite this class imbalance
the median fraction of multiple matches correctly predicted
is 
83\% 
on training data and 83\% across the verification datasets
(Figure~\ref{fig:verify_multiple}).

The three trees of the smallest
best of generation 100 program 
have a combined size of 254.
Hence we try to explain each part of it individually.
GP's overall prediction is given by the sign of
Sum1$-2$M$_{36}$.
However
M$_{36}$ tends to be small compared to Sum1.
Thus the sign of Sum1
plays the major role in predicting repeated Human DNA sequences.

Even the simplified prepas1 tree used to calculate Sum1 
has 96 elements.
Each DNA base it forms a non-linear composition of 
data calculated by prepas0, 
data from the last time it was called
and G
(i.e.\ if the current base is a G or not,
\mbox{Section~\ref{sec:bases}}).
Thus 
the prediction depends only on G
and, via prepas0,
Qual and Run.
In fact G only plays a minor role,
suggesting GP has found a way to predict if BLAST will find multiple
high quality matches which 
mostly uses the quality of the Solexa scan
and BLAST's dislike of simple patterns
(which tend to give Run larger values).

Obviously predicting repeated patterns
in the three billion Human DNA bases
from a sample of just 36 is very hard.
So it is surprising that the evolved predictor can 
find 83\% of those reported by BLAST\@.
However the above analysis hints that GP
is not only using the 36 bases but also
inferring something about the way BLAST deals with 
and reports repeats.

\newpage
\section{Conclusions}

\noindent
Considerable manual effort is needed to create programs. 
In many cases problems do not have clear cut solutions
but instead there may be a range of solutions which make different
compromises between various benefits and differing types of cost.
Where the solutions involve software,
it may be impossible to know how good a trade-off a solution offers
until the software has been coded.
Today even 
identify new operating points 
is often highly labour intensive and few projects can afford to even explore
more than one possibility by hand.
Automated software production
offers the prospect of exploring complete Pareto trade-off surfaces,
for example,
between functionality and speed.

As a prelude to this,
we have explored modelling 
critical aspects of an important non-trivial program (BLAST).
Since in future we hope to consider automatically generating bespoke
software,
we deliberately restricted our analysis to one
of the many applications where the program might be applied.
We chose 
mapping human genetic variation using Next-Gen DNA sequences
from the 1000 genomes project.
This is an important task 
for which
the data are both plentiful and freely available online
and it is well beyond our target program's current abilities.
Our goal is not to make better generic code but to improve existing
code by specialising it to each task.
The importance of the 1000 genomes project 
and the shear volumes of data make this
an ideal candidate for dedicated software.
If software was cheap enough other applications would each have their
own bespoke code.

Although our models were created by considering BLAST's outputs,
BLAST is the de~facto standard for Biological string matching
which other programs seek to emulate.
Thus although our models deliberately apply only
to short (36 bases) human DNA sequences 
generated by the Solexa scanners used by the 1000 genomes project,
they can be thought of as embodying 
to a greater or less extent aspects of an idealised
mapping between this Solexa data and the reference human genome.
Thus they might also apply to other tools
(e.g.\ BWA and Bowtie2)
which emulate BLAST but are much faster.
Our models cannot replace such programs but where data are plentiful
they might be used to prioritise them.
E.g.\ so that DNA sequences which are expected to yield unique high
quality matches are processed first.

One of the critical components of string matching
is the
use of prefect match ``seed'' regions.
These are short 
fixed length sequences which exactly match the database
and so correspond to our length of matching region.
As explained in
Section~\ref{sec:len_results},
correctly choosing the seed length has a considerable impact on performance.
Thus, although our models are not perfect,
being able to predict the length of the region where the string match exactly
would be of great benefit.

This modelling work shows
GP can predict the quality of the match (E)
(Section~\ref{sec:E_results}), 
length of matching sequences
(Section~\ref{sec:len_results}), 
if a sequence will match
(Section~\ref{sec:match_results}), 
and can even find 83\% of multiple matches
(Section~\ref{sec:repeats}).
Analysis of the evolved predictors of
E value and length of match
shows a strong dependence on Solexa's two quality indicators.
Solexa's
Qual also figures in the evolved prediction of whether 
there is a
match or not.
It also depends upon the sequence's entropy.
Whilst other factors seem to relate to quality of the input data,
the use of entropy hints at a degree of reverse engineering of BLAST,
which does not return low entropy sequences.

There are many repeated sequences in the
human genome.
They make it difficult to map genetic differences, such as SNPs,
(one of the main goals of the 1000 genomes project) 
and therefore 
``reads that map to multiple sites in the genome are usually discarded''
\cite[page~1]{Cheung:2011:NAR}.
Not only are such matches useless 
they also slow down search 
and yet GP can predict 83\% of them in advance.

Although it took BLAST 108 hours,
excluding I/O,
to process 31\,000 DNA sequences
(Section~\ref{sec:results}),
the GP predictors take on average 
less than 0.3 seconds. 

\subsection*{Acknowledgements}
I would like to thank
\href{http://www.iop.kcl.ac.uk/staff/profile/default.aspx?go=11857}
{Caroline Johnston},
\href{http://jermdemo.blogspot.com/}
{Jeremy Leipzig}, 
\href{https://github.com/keithj}
{Keith James}, 
\href{http://about.me/pablopareja}
{Pablo Pareja}, 
\href{http://www.blogger.com/profile/13765837643388003852}
{Pierre Lindenbaum}
and 
{Larry Parnell}. 
Funded by EPSRC grant
\href{http://gow.epsrc.ac.uk/NGBOViewGrant.aspx?GrantRef=EP/I033688/1}
{EP/I033688/1}.

\pagebreak[4]

\bibliographystyle{named}


\bibliography{references,gp-bibliography,gecco2009}

\end{document}